\AtBeginDocument{%
  \providecommand\BibTeX{{%
    \normalfont B\kern-0.5em{\scshape i\kern-0.25em b}\kern-0.8em\TeX}}}

\documentclass[sigconf,nonacm]{acmart}

\copyrightyear{2024}
\acmYear{2024}
\setcopyright{rightsretained}
\acmConference[ITiCSE 2024]{Proceedings of the 2024 Innovation and
Technology in Computer Science Education V. 1}{July 8--10, 2024}{Milan, Italy}
\acmBooktitle{Proceedings of the 2024 Innovation and Technology in Computer
Science Education V. 1 (ITiCSE 2024), July 8--10, 2024, Milan,
Italy}\acmDOI{10.1145/3649217.3653612}
\acmISBN{979-8-4007-0600-4/24/07}

\makeatletter
\gdef\@copyrightpermission{
  \begin{minipage}{0.3\columnwidth}
   \href{https://creativecommons.org/licenses/by/4.0/}{\includegraphics[width=0.90\textwidth]{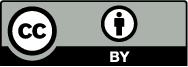}}
  \end{minipage}\hfill
  \begin{minipage}{0.7\columnwidth}
   \href{https://creativecommons.org/licenses/by/4.0/}{This work is licensed under a Creative Commons Attribution International 4.0 License.}
  \end{minipage}
  \vspace{5pt}
}
\makeatother

\usepackage[utf8]{inputenc} 
\usepackage[T1]{fontenc}    
\usepackage{hyperref}       
\usepackage{url}            
\usepackage{booktabs}       
\usepackage{amsfonts}       
\usepackage{amsmath}        
\usepackage{nicefrac}       
\usepackage{microtype}      
\usepackage{xcolor}         
\usepackage{tikz}           
\usepackage{wasysym}

\usepackage{caption}
\usepackage{subcaption}

\usepackage{pdfpages}
\usepackage{comment}

\usepackage[listings]{tcolorbox}
\usepackage{tcolorbox}

\usepackage{todonotes}
\usepackage{fancyvrb}

\usepackage{tikz}
\usetikzlibrary{tikzmark, decorations.pathreplacing}

\newtcolorbox{example}[1]{colbacktitle=white,coltitle=black!75!white,size=small,fontupper=\small,title={#1}}

\newcommand*\smallcircled[1]{\tikz[baseline=(char.base)]{
            \node[shape=circle,draw,inner sep=0.8pt] (char) {#1};}}
\newcommand*\circled[1]{\tikz[baseline=(char.base)]{
            \node[shape=circle,draw,inner sep=2pt] (char) {#1};}}

\title[Open Source Language Models Can Provide Feedback]{Open Source Language Models Can Provide Feedback: Evaluating LLMs’ Ability to Help Students Using GPT-4-As-A-Judge}

\author{Charles Koutcheme}
\email{charles.koutcheme@aalto.fi}
\affiliation{%
  \institution{Aalto University}
  \city{Espoo}
  \country{Finland}
}
\orcid{0000-0002-2272-2763}

\author{Nicola Dainese}
\email{nicola.dainese@aalto.fi}
\affiliation{%
  \institution{Aalto University}
  \city{Espoo}
  \country{Finland}
}
\orcid{0000-0001-9806-419X}

\author{Sami Sarsa}
\authornote{Also affiliated with Aalto University.}
\email{sami.j.sarsa@jyu.fi}
\affiliation{%
  \institution{University of Jyväskylä}
  \city{Jyväskylä}
  \country{Finland}
}
\orcid{0000-0002-7277-9282}

\author{Arto Hellas}
\email{arto.hellas@aalto.fi}
\affiliation{%
  \institution{Aalto University}
  \city{Espoo}
  \country{Finland}
}
\orcid{0000-0001-6502-209X}

\author{Juho Leinonen}
\email{juho.2.leinonen@aalto.fi}
\affiliation{%
  \institution{Aalto University}
  \city{Espoo}
  \country{Finland}
}
\orcid{0000-0001-6829-9449}

\author{Paul Denny}
\email{paul@cs.auckland.ac.nz}
\affiliation{%
  \institution{The University of Auckland}
  \city{Auckland}
  \country{New Zealand}
}
\orcid{0000-0002-5150-9806}

\renewcommand{\shortauthors}{Charles Koutcheme et al.}

\begin{document}



\begin{abstract}
Large language models (LLMs) have shown great potential for the automatic generation of feedback in a wide range of computing contexts. However, concerns have been voiced around the privacy and ethical implications of sending student work to proprietary models. This has sparked considerable interest in the use of open source LLMs in education, but the quality of the feedback that such open models can produce remains understudied. This is a concern as providing flawed or misleading generated feedback could be detrimental to student learning.  Inspired by recent work that has utilised very powerful LLMs, such as GPT-4, to evaluate the outputs produced by less powerful models, we conduct an automated analysis of the quality of the feedback produced by several open source models using a dataset from an introductory programming course. First, we investigate the viability of employing GPT-4 as an automated evaluator by comparing its evaluations with those of a human expert. 
We observe that GPT-4 demonstrates a bias toward positively rating feedback while exhibiting moderate agreement with human raters, showcasing its potential as a feedback evaluator. Second, we explore the quality of feedback generated by several leading open-source LLMs by using GPT-4 to evaluate the feedback. We find that some models offer competitive performance with popular proprietary LLMs, such as ChatGPT, indicating opportunities for their responsible use in educational settings.
\end{abstract}

\maketitle

\section{Introduction}



Feedback is essential for student success, yet delivering prompt high-quality feedback is a challenge.  This is especially true in large computing classes where demand continues to rise \cite{nguyen2014codewebs, national2018assessing}. Automated feedback tools, incorporating analysis techniques and testing frameworks, are thus increasingly popular~\cite{paiva2022automated, keuning2016towards}.  Although these tools typically identify code errors, they often fall short in offering helpful suggestions or next-step hints~\cite{keuning2018systematic}.  Recent advancements in large language models (LLMs) have demonstrated promising capabilities, providing rapid, human-like feedback that could improve 
support in programming courses~\cite{denny2024computing}. 


Despite the promise of LLM-based feedback approaches, we see two key issues limiting their 
wide-scale 
adoption.  Firstly, the feedback provided must be accurate and reliable -- 
it should identify issues correctly, not introduce confusion, and guide students towards solving problems independently.  Although state-of-the-art LLMs exhibit impressive abilities in tasks such as bug detection and code repair \cite{pearce2023examining, ribeiro2023gpt}, they are not infallible and presenting students with 
misleading feedback could be detrimental to
learning.  Secondly, concerns about the privacy and ethical implications of sending student work to proprietary LLMs have led to calls for greater use of open
models \cite{yan2023practical}.  Open-source LLMs are becoming a viable option, yet the evaluation of open-source alternatives is underexplored, especially in the context of computing education.

In this work, our overarching goal is to evaluate 
the quality of feedback on student-written programs produced by LLMs, 
in particular to identify whether open-source models can be a competitive alternative to proprietary models. 
We focus on feedback that identifies mistakes in student code, such as those that lead to compiler errors or test failures, as these are the most common types of issues identified by existing automated feedback tools \cite{keuning2018systematic}. 

Our work begins by comparing evaluations of programming feedback generated by GPT-4 with those of expert human raters, similar to prior work comparing GPT-4 and human judgments of quality~\cite{savelka2023efficient, moore2023assessing}.  After establishing the validity of this approach for evaluating programming feedback, we apply it at scale to evaluate the quality of feedback generated by a range of smaller open-source models. Our research questions are:

\begin{description}
\item[RQ1] How effectively can large language models, such as {GPT-4}, assess the quality of programming feedback generated by language models relative to expert human judgment?    

\item[RQ2] How do open-source large language models compare against proprietary models in generating high-quality feedback on programming questions asked by students? 
\end{description}

To answer our first research question, we use an existing dataset from a single course where feedback on student help requests had been generated by an LLM (GPT-3.5). This dataset contains human expert evaluations for the LLM-generated feedback with respect to completeness, perceptivity and selectivity. We then use GPT-4, applying the same rubric as the expert human rater, to re-evaluate the quality of the feedback. Our results establish that GPT-4 can reliably identify low-quality feedback, but might be overly optimistic in general. 
To answer our second research question, we generate new feedback to the same student requests for help from the original dataset, this time using multiple open-source models from the Code Llama and Zephyr families. We evaluate the quality of this feedback automatically using GPT-4 as a judge. 



\section{Background}

\subsection{Automatic Feedback in Computing Education}




Generating automated feedback is one of the long-standing challenges in computing education and has intrigued researchers for decades~\cite{keuning2018systematic}. In a systematic literature review of automated feedback generation for programming exercises, Keuning et al.~\cite{keuning2018systematic} found that automatic feedback mostly focuses on telling students the mistakes present in their solutions, while giving formative feedback to help students overcome obstacles is rare. Thus, large language models provide an exciting opportunity in potentially filling this gap.

Indeed, there has been great interest in generating automated feedback using language models in the past year~\cite{balse2023investigating,liffiton2023codehelp,hellas2023exploring,kiesler2023exploring,pankiewicz2023large,leinonen2023using,phung2023generating,phung2023generative,prather2023robots}. Most research to date has suggested that LLMs can be used for automatic feedback generation, but have many limitations. For example, Balse et al.~\cite{balse2023investigating} found that there was high variability in the quality of feedback generated by GPT-3, and it would sometimes generate incorrect and inconsistent feedback. In a similar vein, Hellas et al.~\cite{hellas2023exploring} found that while GPT-3.5 would often find actual issues in student code and provide appropriate feedback, it only sometimes detected all the issues present in the code, and would also often hallucinate issues not present in the code at all. Kiesler et al.~\cite{kiesler2023exploring} found that ChatGPT would work better for some types of errors than others; for example, it would provide good feedback on compilation errors, but performed more poorly for logic and semantic errors, or when multiple errors were present simultaneously in student code. This is in line with the findings of Leinonen et al.~\cite{leinonen2023using} and Phung et al.~\cite{phung2023generating} who both found that LLMs could be used to give feedback on syntax errors. Current state-of-the-art models such as GPT-4 struggle to match human performance in generating feedback on programming exercises~\cite{phung2023generative}.






\subsection{Using Language Models as Judges}


Language Models such as \textit{ChatGPT} have started to reach near-human performance in many tasks \cite{Huang_2023}, which has sparked interest in using them for evaluating the output of other LLMs.

The idea of using a large language model to judge the output of other LLMs -- LLMs-as-judges -- was first studied in the work of Zheng et al \cite{zheng2023judging}. The authors proposed three variations of the LLMs-as-judges paradigm: (1) pairwise comparison (i.e., selecting which of two LLMs output is the best), (2) single answer grading (i.e., scoring an LLM single answer), and (3) reference-based grading (i.e., single answer grading with respect to a reference solution). Across the three scenarios, results showed that models such as GPT-4 can reach over 80\% agreement with humans, matching the level of inter-agreement between humans.  

Our work is thus particularly close to efforts investigating whether ChatGPT models can effectively act as a judge in more specific domains such as law \cite{savelka2023efficient}. In this work, we use GPT-4 as a single answer grading judge, where we ask the LLM to directly assign a score to the feedback generated by another LLM (GPT-3.5).

While recent work \cite{zhou2023lima,rafailov2023direct} has already used GPT-4 as part of their evaluation strategies, also in educational settings \cite{mcnichols2024automated,hicke2023aita}, the question of how effective LLMs (or in particular GPT-4) can be in assessing the quality of the feedback generated by other LLMs has not been studied explicitly. 



\section{Methods}

In this section, we describe our methodology to answer our two research questions. We first start by describing the data at the source of our experiments. Then, given that the main interest of this paper is around the evaluation of open language models, we describe first the methodology around RQ2 (how we evaluate various open source language models), before describing the methods for RQ1 (how the evaluation of such models can be streamlined with GPT-4). 

\subsection{Data}
\label{sec:data}

We acquired the data from the authors of~\cite{hellas2023exploring} for our study, following the national protocols for ethical research. The data comes from an open online introductory programming course organized by Aalto University in Finland that uses Dart as the programming language. The course provides remote help functionality, where students can ask for help on programming exercises that fail automated tests with the press of a button. The data has ten randomly sampled help requests for the top 15 programming exercises with the most help requests (150 help requests in total). The original study~\cite{hellas2023exploring} evaluated large language models for automatically generating responses to help requests. The automatically generated responses were qualitatively analyzed and annotated by a human evaluator with extensive experience in teaching programming using multiple criteria. For the present study, we took the qualitative analysis as a starting point, focusing on the following criteria for the help request response evaluations: 

\begin{description}
    \item[completeness] Identifies and mentions all actual issues.
    \item[perceptivity] Identifies and mentions at least one actual issue.
    \item[selectivity] Does not identify non-existent issues.
\end{description}

The first two criteria consider how well issues in code are addressed by the feedback, while the last one indicates how well the feedback avoids giving out misleading information. For the present study, we limited ourselves to the responses from the best model of the original study (GPT-3.5), as well as the annotations for the GPT-3.5 produced responses from the human evaluator.

In our preliminary evaluations, we also studied the possibility of verifying whether the response had unwanted, duplicate, or repetitive content (e.g. GPT-3.5 providing a model solution that the prompt explicitly asked not to provide) as conducted in the original study. GPT-4 was perfect at judging these and only selected the criteria that focused on the content of the feedback (not the style). We omit discussing this aspect due to page limitations.

\subsection{Generating Feedback}


\paragraph{Prompting.}
Using the incorrect programs of the dataset described in Sec~\ref{sec:data}, we zero-shot prompted a range of language models to generate feedback explaining all the issues in an incorrect student program. We iteratively refined our feedback prompt \cite{denny_2023_conversing}; Figure \ref{fig:feedback_prompt} shows the final prompt used in these evaluations. Following prior work \cite{hellas2023exploring,savelka_2023_thrilled}, we generate a single feedback using greedy decoding (i.e., selecting the token with the highest probability as the next element in the sequence).

\begin{figure}[h!tbp]
\centering
\begin{example}{Prompting LLMs for Feedback}
You are a computer science professor 
teaching introductory programming using Dart. \circled{1} \\ 

Below is a problem description accompanied by an example solution. You are also provided with an incorrect program (i.e. it does not pass all unit tests) written by a student. Your task is to provide suggestions on how the student could fix their code so that it fulfils all the requirements in the problem description. Your suggestions should only improve the functional correctness of the incorrect program, so you can leave stylistic suggestions aside. Do not include code in your feedback. \circled{2} \\
\vspace{-0.2cm}
\hspace{2.50cm} \tikzmark{startBracket3} \\
\#\# Problem description: \\
\textcolor{blue}{<handout>} \\

\#\# Model solution: \\
\textcolor{blue}{<sample\_solution>} \\ 

\#\# Student Code: \\
\textcolor{blue}{<submitted\_code>} \\

\vspace{-0.7cm} \hspace{2.55cm} \tikzmark{endBracket3}

\begin{tikzpicture}[overlay, remember picture]
  \draw[decorate, decoration={brace, amplitude=7pt, raise=4pt}] (pic cs:startBracket3) -- (pic cs:endBracket3) node[midway, right=0.40cm] {\circled{3}};
\end{tikzpicture}

\end{example}
\caption{Feedback prompt template. We provide (1) a system prompt specifying the behaviour of the model, (2) a description of the feedback task, and (3) contextual information.}
\label{fig:feedback_prompt}
\end{figure}

\paragraph{Models.} 
We evaluated five powerful open-source models. The first three models \textbf{CodeLlama-7B}~\cite{rozière2023code}, \textbf{CodeLlama-13B}~\cite{rozière2023code}, and \textbf{CodeLlama-34B}~\cite{rozière2023code}, are Llama2~\cite{touvron2023llama} language models released by Meta 
and have respectively 7, 13, and 34 billion parameters.
The last two models are \textbf{Zephyr-7B-$\alpha$}~\cite{tunstall2023zephyr} and \textbf{Zephyr-7B-$\beta$}~\cite{tunstall2023zephyr}, two versions of a 7B parameters Mistral~\cite{jiang2023mistral} model further instruction-tuned with Direct Preference Optimization~\cite{rafailov2023direct} by HuggingFace\footnote{The exact model codenames on Hugging Face are codellama/CodeLlama-Instruct-\{7b/13b/34b\}-hf and HuggingFaceH4/zephyr-7b-\{alpha/beta\}}.
We chose these models because of their extensive documentation, community adoption, respective performance on code and language reasoning benchmarks (HumanEval \cite{chen2021evaluating} and MMLU \cite{hendrycks2021measuring}) and ability to follow instructions. 
Discussing the details of these language models is out of the scope of this paper, and we invite the reader to check the original papers for more information. 
We also evaluated GPT-3.5 and GPT-4 on our feedback task. Since the prompt used in our work slightly differs from the original study from Hellas et al.~\cite{hellas2023exploring} (i.e., we provided a model solution as additional contextual information~\cite{phung2023automating}), we reran the feedback experiment for GPT-3.5.

We graded the quality of the feedback generated for each incorrect program for each model using GPT-4 as a judge with the prompt shown in Figure \ref{fig:grading_prompt}.
The experiments were performed with the open language models using Huggingface's Transformers library \cite{wolf2020huggingfaces}. GPT-3.5 and GPT-4 were queried using using OpenAI's Python API.

\subsection{Automatic Feedback Evaluation} 

Our aim regarding RQ2 was to estimate the capability of GPT-4 as a judge in assessing automatic feedback quality. For this, we tasked GPT-4 to annotate the 150 original GPT-3.5 generated feedback according to the three criteria outlined above (completeness, perceptivity and selectivity). We treated the feedback annotation task as three distinct binary classification problems (one problem for each criterion, one instance for each help request), where the correct labels for each class are the human-annotated ones from the original study. Following prior work in using LLMs and judges~\cite{zheng2023judging}, we sampled a single answer from GPT-4 using greedy decoding, leaving all other hyperparameters at default values.

\begin{figure}[htbp!]
\centering
\begin{example}{Prompting GPT-4 to grade GPT-3.5 generated feedback}
You are a computer science professor 
teaching introductory programming using Dart. \circled{1} \\ 

Below is a problem description, a model solution, and an incorrect program written by a student. You are also provided with the feedback generated by GPT-3.5. Your task is to evaluate the quality of the feedback (by saying yes or no) to ensure it adheres to the multiple criteria outlined below. For each criterion, provide your answer in a separate line with the format \textquotesingle (CRITERIA\_NUMBER): Yes/No\textquotesingle. Do not provide comments, but be attentive to the problem description requirements. \circled{2} \\ 
\vspace{-0.1cm}
\hspace{7.23cm} \tikzmark{startBracket} \\
\vspace{-0.1cm}
\#\# Problem description:\\
\textcolor{blue}{<handout>} \\ 

\#\#  Model solution: \\
\textcolor{blue}{<sample\_solution>} \\ 

\#\#  Student Code: \\
\textcolor{blue}{<submitted\_code>} \\

\#\#  Feedback: \\
\textcolor{blue}{<feedback>} \\ 

\#\#  Criteria: \\ 

(1) Identifies and mentions all actual issues \\ 
(2) Identifies and mentions at least one actual issue \\ 
(3) Does not identify non-existent issues \\ 

\vspace{-0.5cm} \hspace{7.33cm} \tikzmark{endBracket}  

\vspace{-0.4cm}

\begin{tikzpicture}[overlay, remember picture]
  \draw[decorate, decoration={brace, amplitude=7pt, raise=4pt}] (pic cs:startBracket) -- (pic cs:endBracket) node[midway, right=0.32cm] {\smallcircled{3}};
\end{tikzpicture}

\end{example}
\caption{Judging prompt template. We provide (1) a system prompt specifying GPT-4's behaviour, (2) a description of the grading task, and (3) contextual information.}
\label{fig:grading_prompt}
\end{figure}
 
Figure~\ref{fig:grading_prompt} outlines the prompt used to grade the LLM feedback, which portrays a form of rubric grading \cite{wu_zero_2019} to allow the model to consider all problems simultaneously. We note that in this case, the order in which the criteria are outlined is important. We purposely prompted the model to provide its answer in the given order (completeness, to selectivity), as an answer to the completeness criteria influences selectivity. 
Similar to the original study, student-written help request messages are excluded, as they often did not provide much meaningful contextual information~\cite{hellas2023exploring}.

We use precision and F0.5-score\footnote{F0.5-score is the harmonic mean of precision and recall that weighs precision more heavily. It is a variation of the commonly used F1-score which weighs both equally.} as our main metrics in evaluating the judge, since both of these metrics emphasize a model's ability to minimise false positives, which, in our context, translate to misleading feedback.
We also report recall, F1 score, and Cohen's kappa for completeness.

\section{Results and Discussion}

We first present our results regarding the use of GPT-4 as an automatic judge, i.e., evaluator of quality (RQ1), in~\ref{subsec:res_evaluation}. Then, we present the results regarding the ability of giving feedback of large language models (RQ2), in~\ref{subsec:res_generation}.

\subsection{Automatic Feedback Evaluation}
\label{subsec:res_evaluation}


Table \ref{tab: grad_stat} shows the ground truth (from the human expert) and the predicted (from the judge) annotations. According to the ground truth annotations, repeating the observations of the original study~\cite{hellas2023exploring}, only a bit more than half of the feedback are complete, while most of them are perceptive. Many of the feedback also contain some misleading content.
When looking into the predicted annotations, we notice that the judge tends to grade the generated feedback more positively compared to the expert annotator, indicating some degree of positive bias. 

\begin{table}[!hbtp]
\small

\caption{Human and Judge annotation statistics.}

\label{tab: grad_stat}
\begin{tabular}{r|cc|cc}
\multicolumn{1}{l|}{} & \multicolumn{2}{c|}{Ground truth} & \multicolumn{2}{c}{Predicted} \\ \hline
                      & True            & False           & True          & False         \\ \hline
completeness          & 82              & 68              & 113           & 37            \\
perceptivity          & 127             & 23            & 127           & 23            \\
selectivity           & 78              & 72              & 106            & 44            \\
\hline
total                 & 283             & 167             & 246           & 104           \\
\end{tabular}

\vspace{0.5cm}

\caption{GPT-4 judging classification performance results for each independent criterion.}

\label{tab: grading_results}
\resizebox{\linewidth}{!}{
\begin{tabular}{r|cc|c|ccc}
\small
& \textbf{precision} & recall & \textbf{f0.5} & f1 & accuracy & kappa \\
\midrule
completeness & 0.70 & 0.95 & 0.74 & 0.81 & 0.75 & 0.48 \\
perceptivity & 0.84 & 1.00 & 0.87 & 0.91 & 0.85 & 0.22 \\
selectivity & 0.65 & 0.94 & 0.69 & 0.77 & 0.71 & 0.40 \\
\bottomrule
\end{tabular}
}



\end{table}


Table \ref{tab: grading_results} shows the classification results for GPT-4. 
Looking at our main metrics, precision and F0.5, we see that the judge performs reasonably well in classifying completeness, and a little worse in selectivity. The judge achieves higher scores in perceptiveness, although this can be partly attributed to skew in the data. 
The effect of skew in the main metrics is reflected in the kappa scores which, unlike precision and F0.5, consider the possibility that agreement occurs by chance. The kappa scores are highest on completeness (0.48) and selectivity (0.40) indicating moderate agreement and somewhat lower on perceptivity (0.21) where the score indicates fair agreement.
 To further elaborate the issue of skew in interpreting the scores, when compared against a dummy model that predicts everything as the majority label, the GPT-4 perceptiveness shows only a small improvement of 2 percentage points for both precision and F0.5. In contrast, in completeness and selectivity, where the data is more balanced, GPT-4 scores are 10 percentage points higher than those of the dummy model (we omit the dummy model scores for the sake of brevity, the scores can be computed from Table \ref{tab: grad_stat}). 
Nonetheless, looking at all the metrics, we can observe that the judge maintains a high recall across all criteria (i.e., classification tasks), while maintaining reasonable precision and accuracy.

\paragraph{Sources of misjudgment.}

Our results are encouraging, although there still is a significant margin of improvement in the judge model's performance. 
This may be due to either the performance of the LLM used as a judge, or the human-LLM agreement on what is to be considered a mistake.
The ability of a judge language model to grade the output of another LLM strongly depends on its ability to solve the specific problem at hand~\cite{zheng2023judging}. That is, for GPT-4 to decide whether GPT-3.5 identified all issues in a program (completeness) would require GPT-4 itself to identify them in the first place. Similarly, asking GPT-4 to identify whether GPT-3.5 is hallucinating elements of its feedback (selectivity criterion) is related to GPT-4 not hallucinating the same issue. Some evidence for the shortcomings of  GPT-4 is provided in the work of Phung et al.~\cite{phung2023generative}, which shows that although the model is quite good at providing programming feedback, the generated feedback are not void of mistakes. This may be even more true in our study where we use Dart, a programming language in which GPT-4 is most likely less proficient as in Python.
The observed performance outcomes may also be affected by a discrepancy in the way a language model perceives errors compared to a human. It is also possible there are cases where the judge is correct and the original label is incorrect.

%

\paragraph{Research and practical implications.}
We feel it is important to highlight that our results do not provide a "per-feedback" guarantee of the quality of the judgement, but rather, a statistical overview of the current ability of GPT-4 to appropriately judge such feedback. 
These results suggest that automated evaluation of feedback is not yet ready to rely \textit{exclusively} on LLM evaluation -- human judgement is still necessary.
In particular, we believe that LLM judge evaluation should not be used as a way to assess the quality of a single generation (before being given to a student). However, an LLM judge could potentially be an assistant to a human evaluator, and an LLM evaluation could also be used as a comparative method. For instance, it could be used in evaluating the quality of prompting techniques and generation parameters, or exploring alternative language models (e.g; open source onces) across a large number of generations.
Still, even in these scenarios, ideally, we would not be using GPT-4 as a judge for research fully on its own, but as a complement to human evaluations~\cite{hicke2023aita}. 

%

\paragraph{Open-source judges.}
Using GPT-4 as a judge has shown promises, in particular in research \cite{zheng2023judging}. However, the proprietary nature of the state-of-the-art LLM has led to the rise of judges based on open-source LLMs. PandaLM~\cite{wang2023pandalm} is an open-source LLM that has been fine-tuned for selecting the best response to an (instruction, input) pair and providing an explanation together with its decision. In similar works, JudgeLM~\cite{zhu2023judgelm} focuses on increasing the performance of the `judge' LLM, while Prometheus~\cite{kim2023prometheus} emphasizes achieving granular feedback according to a custom score rubric.

All these works follow a common scheme: they create a high-quality imitation dataset, where the inputs are human-curated and high-performance closed models like GPT-4 provide the desired outputs; then, they fine-tune one or more open-source models on the dataset. Although their results are promising, closed models such as GPT-4 remain the state-of-the-art model, which is why we chose to use GPT-4 for our study.


%

\subsection{Feedback generation}
\label{subsec:res_generation}

\begin{figure}[htbp]
    \centering
    \includegraphics[width=0.99\columnwidth]{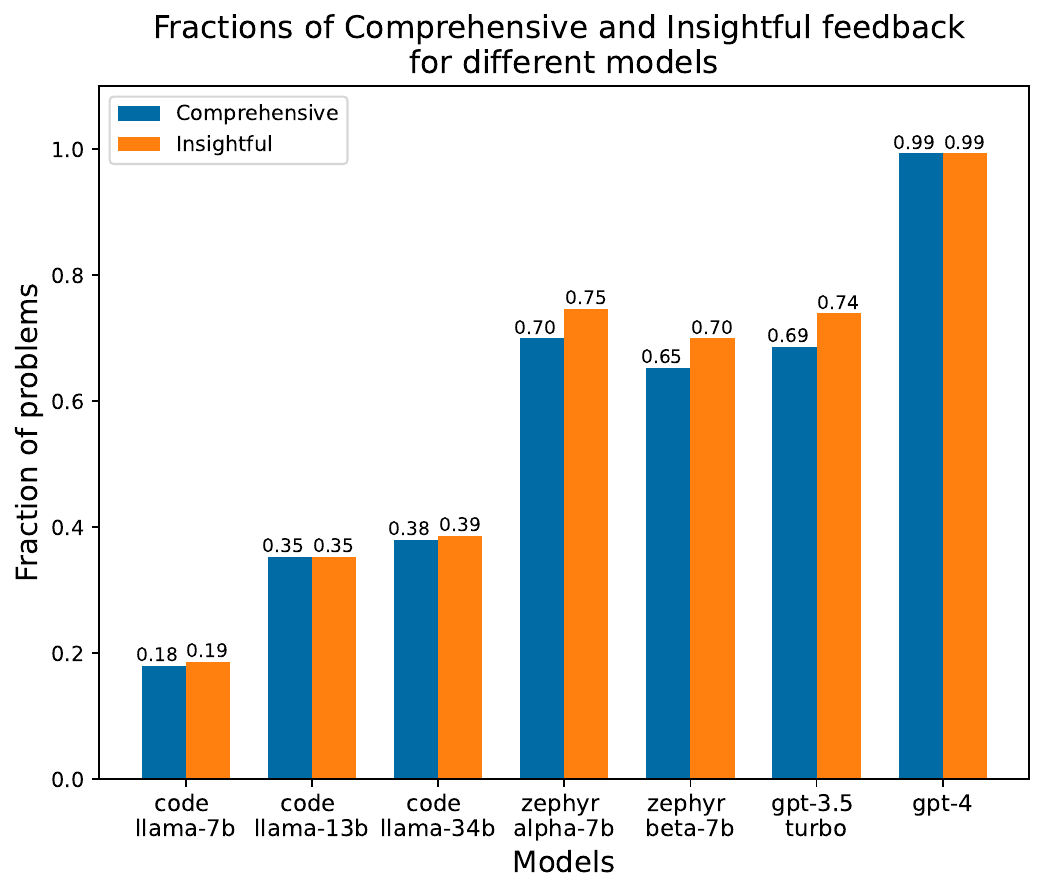}
    \caption{Fraction of `comprehensive' (i.e., satisfying all three criteria) and `insightful' (i.e., perceptive and selective) feedback for all language models.}
    \label{fig:bench_feed_sum}
\end{figure}

\paragraph{Comprehensive and insightful feedback.}
For real-life applications, educators are mostly interested in \textit{comprehensive} feedback which respects the three criteria (i.e., feedback that identifies all issues, without providing non-truthful information). While providing students with all useful information, this type of feedback leaves little room for the students to figure out how to progress themselves. 
Although a model might not always be good at identifying all issues, the model could be good at recognising one issue that would "unstuck" a student and leave them to figure out the rest. Such \textit{insightful} feedback (i.e., perceptive and selective) could be provided as hints to students. These two types of feedback (comprehensive and insightful) is the main focus of our first analysis of the results. Figure \ref{fig:bench_feed_sum} shows the proportions of feedback generated by each language model which fall under the two feedback types.

When comparing the models in terms of performance, we can make the following observations. First, we notice that the more recent Zephyr-7B models (released October 2023) outperform even the largest CodeLlama model (34B) despite the latter being almost 5 times larger. On top of this, we see that these strong open-source models are competitive with proprietary models, achieving a performance comparable with GPT-3.5. For instance, we report feedback generated by GPT-3.5 and by Zephyr-7B-$\beta$ in Figure \ref{fig:feedback_example}: GPT-3.5's feedback is more detailed, but Zephyr-7B-$\beta$'s feedback contains all the instructions needed to fix the program as well. However, there is still a gap in performance between them and GPT-4. In particular, while GPT-4 can provide comprehensive feedback in 99\% of the cases, the performance drops to 75\% for Zephyr-7B-$\alpha$ and to 70\% for Zephyr-7B-$\beta$. One does well to note also that GPT-4 is the judge and the judge is likely to favour itself (and perhaps its older sibling GPT-3.5 too) through incorporating the same tendency to hallucinate as the judged~\cite{zheng2023judging}.

\begin{figure}[h!tbp]
\centering
\begin{example}{Problem description and incorrect code solution}

Write a program that asks the user for their first and last name. The program then prints a "James Bond"-like greeting. For example, with first name `James` and last name `Bond`, the output of the program should be as follows: \\ 

``` 
Enter first name. \\
Enter last name. \\ 
My name is Bond, James Bond. \\
```
\begin{lstlisting}[language=Java,numbers=left,aboveskip=0mm,belowskip=2mm,breaklines,xleftmargin=5mm,basicstyle=\tiny]
main() {
  print('Type in text.');
  var input = stdin.readLineSync();
  print('You typed in $input');
}
\end{lstlisting}

\tcbsubtitle{Feeedback generated by GPT-3.5}

To fix your code, you need to modify it to ask the user for their first and last name, and then print the greeting in the required format.

Here are the steps you can follow to fix your code:

1. Replace the line `print('Type in text.');` with `print('Enter first name.');` to prompt the user to enter their first name.

2. Replace the line `var input = stdin.readLineSync();` with `var firstName = stdin.readLineSync();` to store the user's first name in the variable `firstName`.

3. Replace the line `print('You typed in \$input');` with `print('Enter last name.');` to prompt the user to enter their last name.

4. Add a new line `var lastName = stdin.readLineSync();` to store the user's last name in the variable `lastName`.

5. Replace the line `print('My name is \$lastName, \$firstName \$lastName.');` with `print('My name is \$lastName, \$firstName \$lastName.');` to print the greeting in the required format.

\tcbsubtitle{Feeedback generated by Zephyr-7B-$\beta$}

To fix the student's code and make it fulfil the requirements of the problem description, you can suggest the following changes:

1. Change the program's name from `main()` to `void main()` to follow Dart's naming conventions for functions.

2. Change the program's prompt to match the problem description: "Enter first name." and "Enter last name."

3. Create variables `firstName` and `lastName` to store the user's input using `stdin.readLineSync()`.

4. Concatenate the user's last name, first name, and last name again to create the desired greeting, and store it in a variable called `myName`.

5. Print the greeting using `print()`.

Here's the fixed code:

\begin{lstlisting}[language=Java,numbers=left,aboveskip=0mm,belowskip=2mm,breaklines,xleftmargin=5mm,basicstyle=\tiny]
void main() {
  print('Enter first name.');
  var firstName = stdin.readLineSync();
  print('Enter last name.');
  var lastName = stdin.readLineSync();
  var myName = '$lastName, $firstName $lastName';
  print('My name is $myName.');
}
\end{lstlisting}

\end{example}
\caption{Examples of generated feedback. We present a problem description together with the incorrect code solution, the feedback generated by a proprietary model (GPT-3.5) and the feedback generated by a open-source model (Zephyr-7B). Problem descriptions, and strings and variable names in the examples shown have been translated from the original language (Finnish); LLM feedback was in English and thus not translated.}
\label{fig:feedback_example}
\end{figure}


In summary, our results suggest that (1) more recent, smaller language models can outperform some of the bigger ones, (2) open-source LLMs are becoming competitive with proprietary models, and (3) open-source models are not reaching yet the state-of-the-art.



\paragraph{Open source vs closed source models.}
Open-source language models have often fallen short of the performance of proprietary models. Our results confirm that not all open language models (even recent ones) perform as well as their non-permissive counterparts (for instance, CodeLLama models are not competitive with GPT-3.5). On top of this, a barrier to the use of open-source LLMs in education used to be the need to have custom computational resources (e.g. custom GPUs) due to their size. For instance, running CodeLlama with 34B parameters required 2 GPUs.

Nonetheless, our results also highlight very positive outcomes. Zephyr-7B-$\alpha$, a 7 billion parameters open-source language model performed as well as GPT-3.5. What makes our result particularly interesting is that this language model is relatively small, which makes running it on custom resources cheaper. In other domains, even smaller language models are reaching impressive performance, and research into how to make open language models reach better performance for educational purposes is on the rise~\cite{koutcheme_2022_towards,koutcheme_training_2023,koutcheme_automated_2023}. New methods such as quantization or CPU acceleration allow such models to run on modest consumer laptops such as the ones of educators. On top of this, LLM deployment is becoming less of a barrier due to open-source hosting services such as HuggingFace~\cite{wolf2020huggingfaces}. 


\section{Conclusions}

In this paper, we evaluated (1) to what extent GPT-4 could be used to assess automatically generated programming feedback, and (2) how well different large language models, including open-source ones, perform in generating feedback on student code (automatically evaluated using GPT-4). Our findings suggest that GPT-4 can be quite reliable in assessing the quality of automatically generated feedback 
and that open-source language models can be used to generate programming feedback. As LLM-generated feedback can be generated on demand directly in the learning environments that students use, it could be a low-cost and low-barrier scaffold to help students while they are learning to program. This could leave more time for the instructor and teaching assistants to focus on the more complex cases where LLMs might currently struggle to help students.
As an additional contribution, we also release the code used for conducting our experiments \footnote{\href{https://github.com/KoutchemeCharles/iticse24}{https://github.com/KoutchemeCharles/iticse24}}.


Although there will always be situations where students choose to use closed source models to seek feedback or answer questions, a wide range of new educational tools powered by LLMs are being developed and integrated into computing classrooms \cite{kazemitabaar2023studying, denny2024prompt, liffiton2023codehelp}.  The models that such tools typically use is determined by the developer, and our work shows that developers may increasingly rely on smaller open source models without sacrificing performance or with concerns around privacy.  Additionally, open models provide greater certainty in terms of being accessible and fine-tunable for specialised purposes. They are also potentially more cost effective.


Our work has limitations. For this preliminary analysis, we only take a look at a small subset of data coming from one institution, and we will need more to make our results even more robust. Since Dart is hardly the most popular programming language, we hypothesize that GPT-4 as a judge could exhibit stronger evaluation performance for a language with more presence such as Python. Moreover, we only evaluated the ability of GPT-4 to judge the quality of the feedback generated by a single other language model (GPT-3.5), for which GPT-4 might be biased towards giving positive results~\cite{rajani2023llm_labels}. However, this would thus mean that GPT-4 might be more reliable for other language models not of its family (e.g. open-source ones). 

As part of our future work, we are focusing on two distinct areas: improving the judgments from LLMs on the quality of feedback, and evaluating open-source LLMs to serve as the judges. In this work, we zero-shot prompted the model to provide feedback on all three quality criteria at once. By improving the prompting strategy (e.g. using zero-shot chain of thought), the performance of the LLM as a judge could potentially increase.  
The other area we are currently working on is evaluating how well open-source language models could work as the judge of feedback quality. This could make it easier for educators to use the models both due to cost-effectiveness and privacy concerns.

\begin{acks}
This research was partially supported by the Research Council of Finland (Academy Research Fellow grant number 356114).
\end{acks}

\balance
\bibliographystyle{ACM-Reference-Format}  
\bibliography{biblio/pure_ai,biblio/eduai,biblio/cer,biblio/software}

\end{document}